# Clustering large 3D volumes: A sampling-based approach

Thomas Lang

Fraunhofer Institute of Integrated Circuits, Division Development Center X-ray
Technology, Innstraße 43, 94032 Passau, Germany
e-mail: thomas.lang@iis.fraunhofer.de

**Abstract**

In many applications of X-ray computed tomography, an unsupervised segmentation of the reconstructed 3D volumes forms an important step in the image processing chain for further investigation of the digitized object. Therefore, the goal is to train a clustering algorithm on the volume, which produces a voxelwise classification by assigning a cluster index to each voxel. However, clustering methods, e.g., K-Means, typically have an asymptotic polynomial runtime with respect to the dataset size, and thus, these techniques are rarely applicable to large volumes. In this work, we introduce a novel clustering technique based on random sampling, which allows for the voxelwise classification of arbitrarily large volumes. The presented method conducts efficient linear passes over the data to extract a representative random sample of a fixed size on which the classifier can be trained. Then, a final linear pass performs the segmentation and assigns a cluster index to each individual voxel. Quantitative and qualitative evaluations show that excellent results can be achieved even with a very small sample size. Consequently, the unsupervised segmentation by means of clustering becomes feasible for arbitrarily large volumes.

## 1. Introduction

Clustering, i.e., an unsupervised partitioning of a dataset, is an important step in many data processing applications[1], and computed tomography (CT) is no exception. In CT, clustering is mainly used in image segmentation, and there most often in medical imaging[2,3]. A lot of different techniques emerged over the last decades of image processing in computed tomography, including K-Means, DBSCAN, fuzzy clustering, and many more. Most of them retrieve information from a dataset, i.e., they are trained, in order to be applied to and creating a partitioning of unseen datasets. However, while recently sampling theory has been applied in feature space for, e.g., dimensionality reduction[4], little work actually has considered sampling the dataset to make clustering applicable to bigger datasets as they occur in industrial computed tomography. One notable work pursued also a stratified sampling approach to extract a representative subset of the data, for which some (potentially computationally intensive) hashing mechanism defines the stratification on which the dataset is clustered[5]. However, there the datasets consist of about half a million datapoints, whereas in the computed tomography domain one deals with billions of voxels within a single volume.

This work considers large volumetric datasets as they are generated in industrial CT and combines well-known random sampling with the CT domain and clustering therein. Theoretical results will be given, in addition to a practical evaluation.

## 2. Random sampling for large volumes

In order to extract a random sample from a general population of data items, a broad variety of algorithms exist[6]. Most notably there are Bernoulli Sampling, Uniform Sampling, and *Reservoir Sampling*. The latter has many attractive properties, including that one can draw a uniformly distributed sample of a *fixed size* from a potentially *arbitrarily big* population without even knowing its size beforehand. For these reasons, it is especially well-suited to extract a random sample from very large volumetric datasets, which can interpreted as a stream.

*2.1 Reservoir sampling and Algorithm L*

This work specifically focusses on *Reservoir Sampling*, which extracts a uniformly distributed sample from a population in a single linear pass. During that process, the titular reservoir is filled and afterwards randomly selected elements are replaced with new items. After the traversal over the population, that reservoir forms the extracted sample[7]. A later adaption of this algorithm named Algorithm L[8] avoided the costly generation of a random number for each population item by observing that the "jumps" between insertions of items into the reservoir follow a geometric distribution. Thus, Algorithm L, as presented in the following pseudocode, skips over several elements without considering them for insertion, where that gap between insertions is computed by drawing a sample of the geometric distribution. In the following, $rand(a, b)$ draws a sample from the uniform distribution on the interval $[a, b]$, while $randi(a, b)$ draws an integer sample from that discrete range.

**Input**: Population $X = \{x_1, x_2, \dots\}$, sample size $M$
**Output**: Sample $S$
**Algorithm L:**
1. $S = \{x_1, \dots, x_M\}$
2. $W = \exp(\log(rand(0,1))/M$
3. $i = K - 1$
4. **while** not done **do**
    $i \mathrel{+}= \lfloor \log(rand(0,1))/\log(1 - W) \rfloor + 1$
    **if** end of stream not reached **then**
        $r = randi(1, K)$
        $S_r = x_i$
        $W \mathrel{*}= \exp(\log(rand(0,1))/M)$
    **fi**
    **done**
5. **return** S

As motivated, the main benefit of Reservoir Sampling and Algorithm L is that a uniformly distributed random sample of a fixed size $M$ is obtained in a single linear pass over the population: Initially, the reservoir is filled with the first $M$ observed values. Then, each (not skipped) population item at index $i$ is accepted with probability $M/i$ and replaces a random entry of the reservoir. At the end, a uniformly distributed sample is obtained[9].

## 2.2 Stratified random sampling

While Algorithm L is highly efficient in terms of computational effort, it still draws a uniformly distributed sample from the population. However, especially in computed tomography, the grayscale value distributions are often biased towards zero, i.e., a lot more zero values are observed than values characterizing material voxels. Therefore, depending on the dataset, sometimes the aforementioned algorithm does not sample any voxel values from some material because it appears far less often than the value zero. To compensate for this, *stratified* random sampling is preferable.

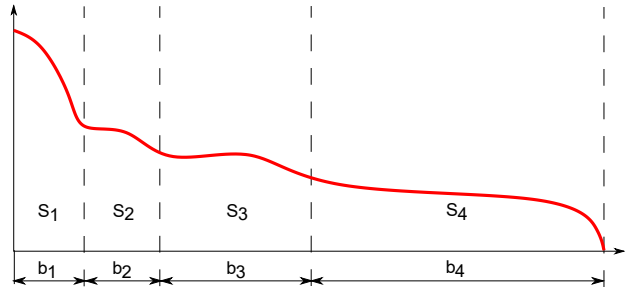

**Figure 1. Illustration of a stratification.**

The key idea is that the range of grayscale values observable in a tomography scan is subdivided a priori into different strati. Afterwards, the above random sampling draws a characteristic subset of values from each stratum, and the overall collection of samples forms the final sample. This idea is illustrated in Figure 1, in which the range of observable grayscale values (the x-axis) is subdivided into four parts to suite the observed grayscale value distribution (the red line). Naturally, an open question is how this subdivision should be done. In principle, any stratification that partitions the range as visualized above is valid. Additionally, in practice the following three strategies were observed to be suitable for X-ray computed tomography given some number $K$ of strati. Here, the stratification strategies partition the unit interval, which then has to be scaled up to fit the range of observable values.

Notationally, expressions of the form $(f(i))_{i=s}^{e}$ denote the ordered tuple of values $(f(s), \dots, f(e))$ and $(a, b) + (c, d) = (a, b, c, d)$ denotes the concatenation of tuples.

### 2.2.1 Linear stratification
Arguably the simplest strategy, it produces evenly spaced partitions, i.e., $B = \left(\frac{i}{M}\right)_{i=0}^{K}$.

### 2.2.2 Exponential stratification
The exponential stratification strategy follows the intuition of the observed grayscale value distributions being heavily biased towards zero and exhibiting an "exponential" form, cf. Figure 1. Specifically, $B = (0,) + \left(2^{1-K+i}\right)_{i=0}^{K-1}$.

### 2.2.3 Mixed stratification
Based on the previous two possibilities, a mixed strategy can be defined which applies an exponential strategy for the first part of the value range (up to some application-dependent threshold) followed by a linear partition of the rest.

*2.2.4 Stratified random sampling*

Combining Algorithm L with some stratification yields the overall stratified random sampling procedure, in which first the stratification is computed, followed by an estimation of the sample sizes per stratum. The latter computes the relative frequencies of values within each stratum in the original volume, and scales it by the overall number of elements that shall be sampled. Next, Algorithm L is executed for each stratum, i.e., a uniform random sample is drawn from the subset of voxels falling into each individual stratum. The concatenation of all samples forms the final extracted sample.

One particular practical issue that needs to be taken care of is the fact, that for some data distributions the sample size in a stratum might be zero. This occurs when the stratum in the overall population contains relatively few values, whose number will be scaled to the potentially small overall sample size. Since samples sizes are always integers, zero sample sizes for individual strati may occur. In such cases, no samples are drawn from these regions, even though the following theoretical results rely on at least one instance being selected from a stratum.

*2.3 Theoretical results*

Concerning Reservoir Sampling, upper bounds for the necessary sample size and the approximation quality of a sample w.r.t. the overall distribution were given in literature. Assuming a population size of $N$ elements and a sample size $M$, denote by $S$ the sample extracted from the population $X$. Then, the relative frequency (which form an approximation of the actual data distribution) approximation is bounded by[10]

$$\sup_{E \in \mathcal{E}} \left| \frac{|S \cap E|}{M} - \frac{|X \cap E|}{N} \right| \leq \mathcal{O}\left( \sqrt{\frac{d + \log(1/\delta)}{M}} \right)$$

with probability at least $1 - \delta$, where $\mathcal{E}$ is a family of subsets of $X$ and $d$ is the Littlestone dimension of $\mathcal{E}$. In simple words, the larger the sample size, the smaller the approximation error, i.e., the better the sample represents the overall population.

The proof of the above statement is a rather technically involved process based on some rather mild assumptions[10]. The reader is encouraged to take a look at this there. In this work, a similar bound for the case when stratification is applied will be given.

**Lemma 1**. Consider the parameters as above and assume the same preconditions[10]. Denote by $c_k$ the relative frequencies of values within stratum $X_k \subseteq X, k = 1, \ldots, K$. Then, the sampling approximation error of stratified Reservoir Sampling is bounded by

$$\sup_{E \in \mathcal{E}} \left| \frac{|S \cap E|}{M} - \frac{|X \cap E|}{N} \right| \leq \mathcal{O}\left( \sqrt{\frac{d + \log(1/\delta)}{M \min_{1 \leq k \leq K} c_k}} \right)$$

with probability at least $1 - \delta$.

*Proof.* By definition, $c_k = N_k/N$ where $N_k$ is the number of items contained in stratum $X_k, k = 1, \ldots, K$. From this, compute the sample size for each stratum as $n_k = Mc_k$. Define $\mathcal{E}_k \subseteq \mathcal{E}$ where $\mathcal{E}_k \cap 2^{X_k} \neq \emptyset$ with $\mathcal{E}$ as above.

First, note that due to pairwise disjointness of the strati

$$|X \cap E| = \left|\bigcup_{k=1}^{K} X_k \cap E\right| = \left|\bigcup_{k=1}^{K} (X_k \cap E)\right| = \sum_{k=1}^{K} |X_k \cap E|$$

and analogously for the samples per stratum $S_k$, for every $E \in \mathcal{E}$. Thus,

$$\sup_{E \in \mathcal{E}} \left|\frac{|S \cap E|}{M} - \frac{|X \cap E|}{N}\right| \leq \sum_{k=1}^{K} \sup_{E \in \mathcal{E}} \left|\frac{|S_k \cap E|}{M} - \frac{|X_k \cap E|}{N}\right|$$

and since the intersections of the strati (and the samples per strati) with population subsets outside of the stratum is the empty set, the right-hand side simplifies to changing $\mathcal{E}$ to $\mathcal{E}_k$. Furthermore, it holds that

$$\left|\frac{|S_k \cap E|}{M} - \frac{|X_k \cap E|}{N}\right| = \left|\frac{|S_k \cap E|}{n_k}c_k - \frac{|X_k \cap E|}{N_k}c_k\right| = |c_k|\left|\frac{|S_k \cap E|}{n_k} - \frac{|X_k \cap E|}{N_k}\right|$$

Consequently,

$$\sup_{E \in \mathcal{E}} \left|\frac{|S \cap E|}{M} - \frac{|X \cap E|}{N}\right|$$

$$\leq \sum_{k=1}^{K} \underbrace{|c_k|}_{\leq 1} \sup_{E \in \mathcal{E}_k} \left|\frac{|S_k \cap E|}{n_k} - \frac{|X_k \cap E|}{N_k}\right| \leq \mathcal{O}\left(\sqrt{\frac{d(\mathcal{E}_k) + \log(1/\delta)}{\min_{1 \leq k \leq K} n_k}}\right)$$

with probability at least $1 - \delta$, where $d(\mathcal{E}_k)$ is the Littlestone dimension of the set $\mathcal{E}_k$. Finally, since for these subsets $d(\mathcal{E}_k) < d$ [10, Eq 15], we obtain

$$\sup_{E \in \mathcal{E}} \left|\frac{|S \cap E|}{M} - \frac{|X \cap E|}{N}\right| \leq \mathcal{O}\left(\sqrt{\frac{d + \log(1/\delta)}{M \min_{1 \leq k \leq K} c_k}}\right)$$

with probability at least $1 - \delta$, which concludes the proof. □

Intuitively, this new bound expresses the same idea that the larger the sample size, the larger the sample sizes for each individual stratum, and thus the better the approximation of the overall data distribution. In this mathematical formulation, the case of empty strati samples is not considered and ill-formed, it is best to handle this programmatically.

# 3. Clustering based on random sampling

The previous algorithms can easily be applied to three-dimensional volumetric datasets as they are frequently generated in computed tomography, by simply interpreting each individual volume element (voxel) with its according grayscale value as a single member of the overall population, which is the volume.
Based on the extracted sample, any application-specific method can be used to gain information or train AI models to perform voxelwise segmentation, for example. Especially in the latter case, this work focusses on clustering algorithms as their training process typically has a polynomial runtime complexity w.r.t. the training data size. On a prior note, it should be mentioned that the following techniques can be applied not only to plain grayscale values but also to feature vectors obtained from the data, by simply changing the models to their multidimensional equivalents. This work, however, focusses on simple clustering based solely on the voxel intensity values.

*3.1 K-Means*

K-Means clustering is perhaps the most popular clustering method following the idea of producing a partitioning $X_1, \ldots, X_K \subseteq X$ of the overall population $X$ by minimizing the distance of each population value to its closest cluster center, i.e.,

$$\operatorname*{argmin}_{X_1,\ldots,X_K} \sum_{j=1}^{K} \sum_{x \in X_j} \|x - \mu_j\|_2^2, \quad \text{s.t.} \quad X = \bigcup_{j=1}^{K} X_j$$

where all subsets $X_j$ are pairwise disjoint and $\mu_j$ denote the mean of each subset.
This problem is known to be NP-hard in general, thus many heuristics exist to solve it reasonably well. One example of which is Lloyd's algorithm[11], which has a polynomial runtime complexity[12].
After training, each voxel is classified by assigning to it the index of the subset whose center is closest to its value, similar to the above formulation. In case of a tie, the smallest index shall be selected.

*3.2 Mini-batch K-Means*

Mini-batch K-Means is one of several adaptions of the classical K-Means clustering algorithm. Specifically, in each iteration of the internal training loop a batch of population items of a configurable size is sampled at random from the population. Based on that subset, the regular K-Means training is performed, and the overall algorithm is continued until some convergence criterion is met[13].
While we refrain from going into more detail here, it is important to note that, contrary to our approach, the internal sampling does neither consider the actual data distribution, nor does it yield any theoretical results. Additionally, it considers the entire population to be present and requires random access, while our method does not. In the experiments conducted, the combination of both methods is studied, i.e., the Mini-batch K-Means algorithm is trained on a random sample extracted by the technique introduced in this work.

*3.3 Clustering using Gaussian Mixture Models*

A popular alternative to classical K-Means clustering is to do the voxelwise assignment using Gaussian Mixture Models, which are probabilistic models that try to approximate the overall grayscale value distribution by a weighted superposition of several Gaussian distributions, i.e., this method assumes a model of the form

$$\sum_{j=1}^{K} \pi_j \, \mathcal{N}\big(\cdot \,\big|\, \mu_j, \Sigma_j\big), \quad \text{where} \quad 0 \leq \pi_j \leq 1, \; \sum_{j=1}^{K} \pi_j = 1$$

In more detail, the goal here is to find the distribution parameters $\mu_j$ and $\Sigma_j$, and also the according weight factors $\pi_j$ of each individual distribution. Typically, an Expectation Maximization algorithm is employed to determine the model parameters, which is guaranteed to converge but might take many iterations. Within each iteration, several matrix multiplications are necessary which scale cubically with the matrix size, hence it is computationally very expensive.

The partitioning is performed by assigning to each population item the index of the Gaussian which maximizes the likelihood function for it. Alternatively, a thresholding method can be derived from such a model for the univariate case[14].

*3.4 Combining clustering with random sampling*

The combination of the introduced random sampling technique and the aforementioned clustering methods is straightforward:

1. Extract a random sample of a fixed size from the volume.
2. Train a voxelwise classifier on the sample.
3. For each voxel in the volume, assign to it a cluster index according to the model.

Since the random sampling only processes local information in linear passes over the volume, and the classification step only concerns local information too, this method is applicable to volumes of **arbitrary** size, as long as the sample fits into memory to train the model. Even better, the overall procedure scales *linearly* with the number of voxels in the volume, while the polynomial runtime for training the model is spent on the sample only, whose size is constant and often independent from the volume size.

On a technical level, the implementation of the random sampling was encapsulated in a C library. Additionally, Python bindings are provided, thus one can easily sample from a stream of numbers (the grayscale values of the data) using the fast library in a setting where the extracted sample is fed to a Python module.

## 4. Results

The experiments conducted in this work are two-fold: Quantitative measurements and qualitative results.

For the quantitative results, a clustering of each dataset using K-Means, Gaussian Mixture Models, and Mini-batch K-Means serves as the baseline. Next, a (stratified) random sample by the proposed scheme was extracted from the datasets and another clustering with these methods is produced. The sample sizes are varied to include small fixed sizes and percentages of the number of voxels of the datum. For each such clustering result, the time required for sampling and for training the clustering method was recorded and compared to the training time from the baselines. Additionally, the baseline and the result based on random-sampling-supported clustering were compared using two clustering similarity metrics (the Fowlkes-Mallows index[15] and the Normalized Mutual Information[16], the latter normalized by the mean entropy). Both metrics compute clustering similarity values between zero (random information, no associations) and one (identical information but probably different cluster indices).

For a qualitative evaluation, Gaussian Mixture Model clustering trained on a stratified random sample (using the exponential stratification scheme) was applied on selected large computed tomography datasets. Note that due to the size of the datasets, an additional quantitative evaluation was not possible in this case.

All results were obtained using the current State-of-the-Art clustering implementation of the aforementioned methods within the Python library *scikit-learn* in combination with random sampling implemented in the programming language C and accessed via accompanying Python bindings.

*4.1 Quantitative Results*

The quantitative analysis was performed for 15 selected computed tomography scans, including industrial objects ("Kolben", "KGH", "Leiterplatte", "Piston", "Tipex", "TPA"), cultural heritage objects ("Tafelklavier", "TRex", "Yorick"), high-resolution scans obtained at a synchrotron beamline ("Bamboo", "Battery", "Brain"), and other scans exposing interesting structural properties ("Ei", "Leberkas", "WurstMitGabel").

Figures 2, 3, and 4 show the collection of all results encoded as a heatmap for Gaussian Mixture Model (GMM), K-Means, and Mini-batch K-Means clustering, respectively. There, the term "simple" stratification corresponds to applying Algorithm L only, i.e., without any stratification. Likewise, Figures 5, 6, and 7 depict the achieved runtime improvements compared to the classical algorithms.

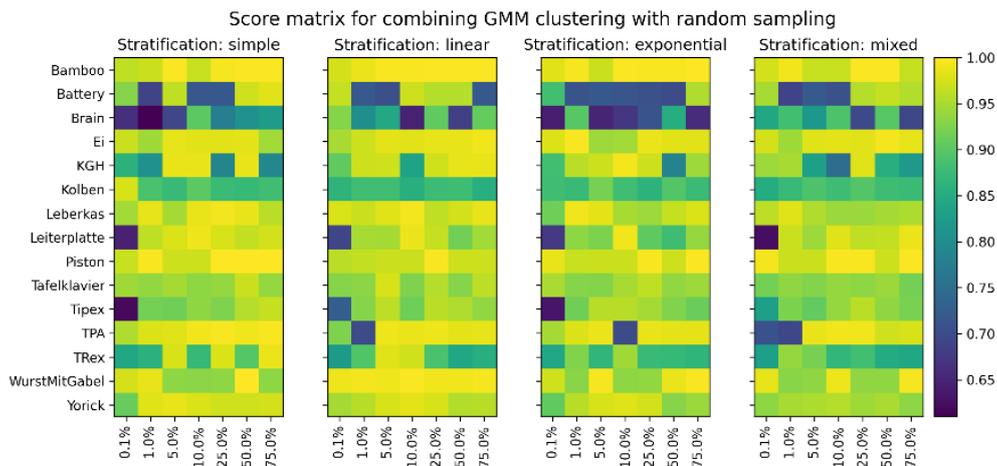

**Figure 2. Clustering scores for Gaussian Mixture Model clustering trained on a stratified random sample over different sample sizes. This is best enjoyed in color.**

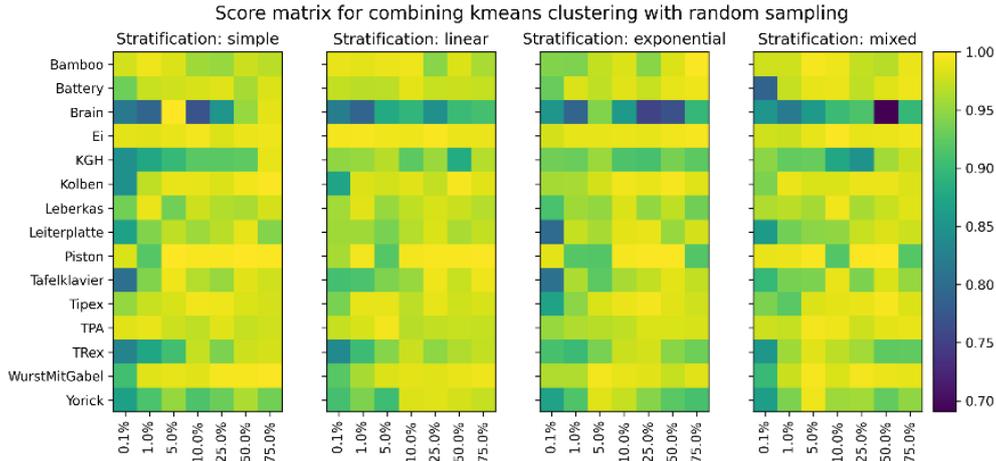

**Figure 3. Clustering scores for K-Means clustering trained on a stratified random sample over different sample sizes. This is best enjoyed in color.**

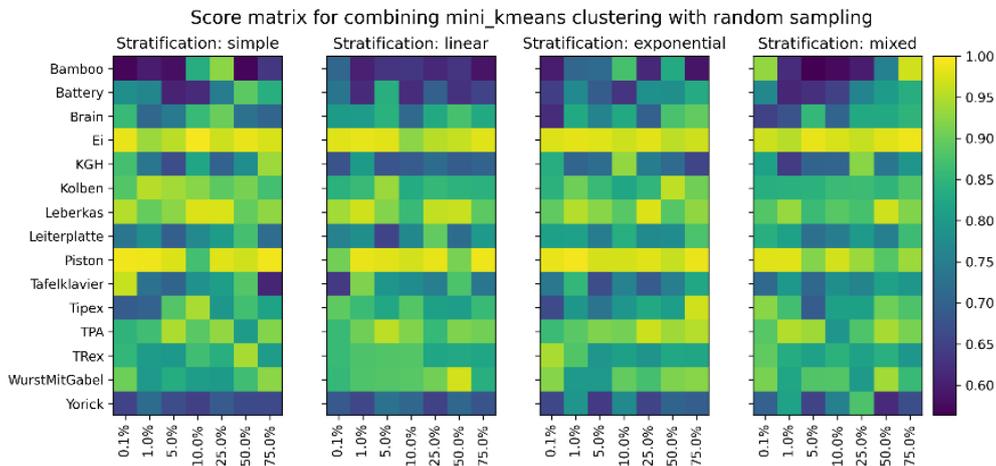

**Figure 4. Clustering scores for Mini-batch K-Means clustering trained on a stratified random sample over different sample sizes. This is best enjoyed in color.**

Generally, good clustering results are obtained even with very small sample sizes of just 0.1 percent of the number of voxels, and mostly the scores increase as the sample size, i.e., the training set size, increases. Naturally, on some scans there is some fluctuation, but it should be noted that the minimal score values are 69%, 61%, and 56% for Gaussian Mixture Models, K-Means and Mini-batch K-Means, respectively. Also, Mini-batch K-Means generally performs not as good when combined with our sampling method. This may be attributed to the fact, that this method already does some internal sampling, thus an additional sampling removes further information. Still, that variant requires random access on the whole dataset, while the proposed sampling scheme does not, which is why the proposed method should be preferred here.

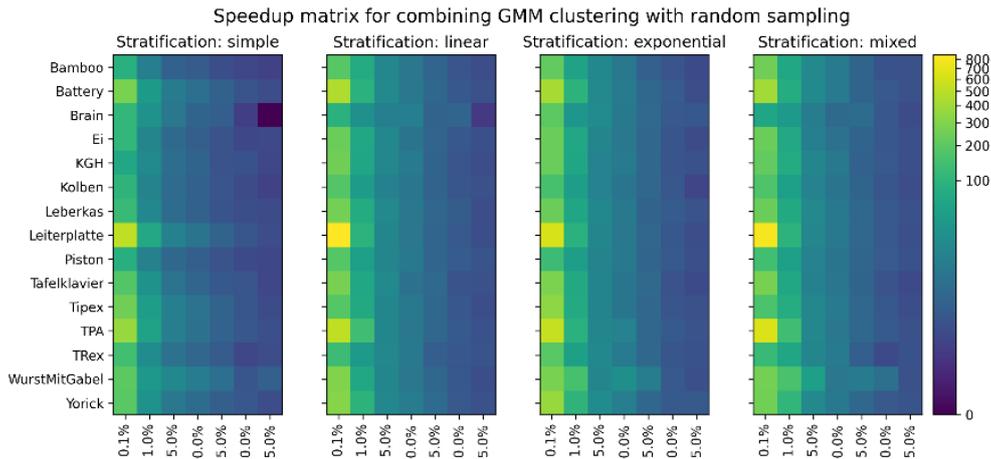

**Figure 5. Speedup over classical Gaussian Mixture Model clustering.**

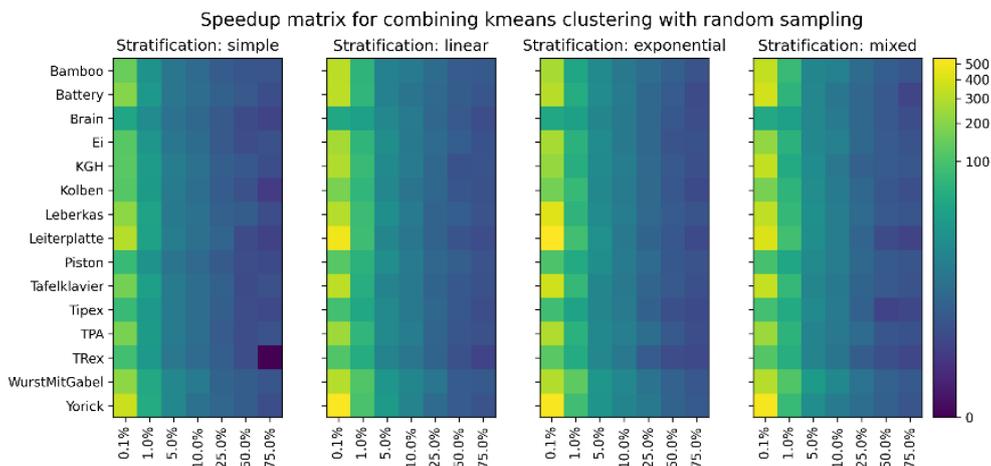

**Figure 6. Speedup over classical K-Means clustering.**

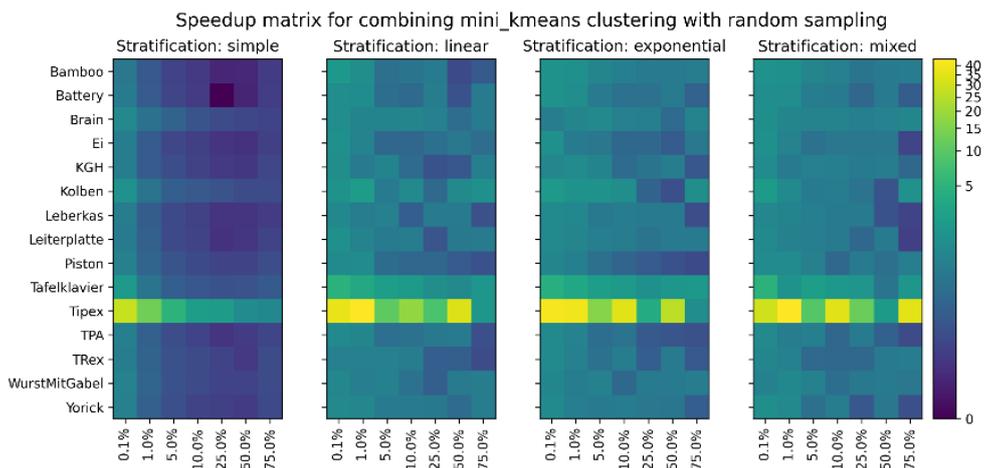

**Figure 7. Speedup over classical Mini-batch K-Means clustering.**

Regarding the speedups, both for clustering using Gaussian Mixture Models and also K-Means huge speedups are achieved (at max here: 851x for GMM, 541x for K-Means), which decrease as the sample sizes increases. The exception to this is, just as before, the Mini-batch K-Means clustering, where in most cases only small speedups or even "speed-downs" occur.

*4.2 Qualitative Results*

Some qualitative clustering results are visualized in Figure 8, showing the application of the clustering method on the scans "Brain" (at 30.80GB), "Kidney" (at 7.26GB), and "Leberkassemmel" (at 4.50GB). The latter was provided by the Fraunhofer Development Center for X-ray Technology, while "Brain" and "Kidney" were taken from the Human Organ Atlas project[17]. For each of these results, the exponential stratification strategy was used to extract a sample of only 4096 elements. The sampling took 77s, 14s, and 9s, respectively. Figure 8 shows the original dataset on the left parts and the clustered results on their right, showing good cluster assignments. Even in noisy and low-contrast scans, the small sample of a size independent of the volume size proved to contain enough information to train a robust model. The proposed procedure was also applied to much larger datasets (currently up to 250GB), which however lack the interesting structural properties of the presented scans.

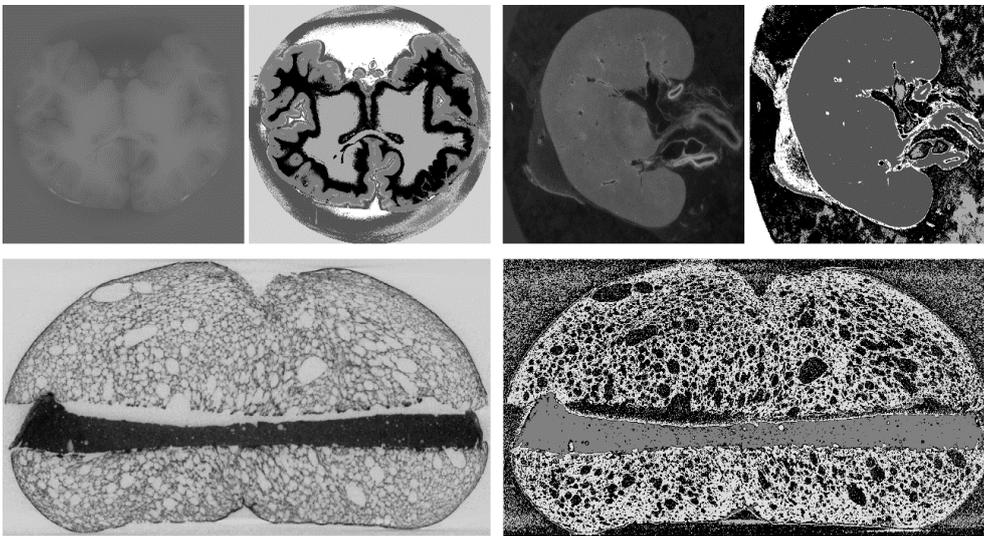

**Figure 8. Qualitative results of "Brain", "Kidney", and "Leberkassemmel".**

## 5. Conclusions

This work proposed the combination of clustering algorithms with random sampling to allow the voxelwise classification of arbitrarily large datasets. A stratification allows incorporating prior information about the grayscale value distribution to enhance robustness. Both a qualitative and a quantitative analysis showed very good results even from very small extracted samples.

**References and footnotes**

1.  A Jain, M Murty and P Flynn, 'Data clustering: a review', ACM Comput. Surv., Vol 31, No 3, pp. 264-323, 1999

2.  M Sutton, J Bezdek, T Cahoon, 'Image Segmentation by Fuzzy Clustering: Methods and Issues', In: Bankman (ed), Handbook of Medical Imaging: Processing and Analysis, pp. 87-106, 2000

3. A Meyer-Baese et al, 'Application of Unsupervised Clustering Methods to Medical Imaging', 5th Workshop on Self-Organizing Maps, 2005

4. N Tremblay and A Loukas, 'Approximating Spectral Clustering via Sampling: A Review', In: F Ros, S Guillaume (eds) Sampling Techniques for Supervised and Unsupervised Tasks, Unsupervised and Semi-Supervised Learning, 2020

5. X Zhao, J Liang and C Dang, 'A stratified sampling based clustering algorithm for large-scale data', Knowl.-Based Syst., Vol 163, pp. 416-428, 2019

6. B Ghojogh et al, 'Sampling Algorithms, from Survey Sampling to Monte Carlo Methods: Tutorial and Literature Review', arXiv, 2020

7. J Vitter, 'Random Sampling with a Reservoir', ACM Trans. Math. Softw., Vol 11, No 1, pp. 37-57, 1985.

8. K Li, 'Reservoir-sampling algorithms of time complexity $O(n(1+\log(N/n)))$', ACM Trans. Math. Softw., Vol 20, No 4, pp. 481-493, 1994.

9. T Roughgarden and G Valiant, 'CS168: The Modern Algorithmic Toolbox, Lecture #13: Sampling and Estimation', Online under http://timroughgarden.org/s17/l/l13.pdf, 2017, accessed on July 7th, 2022

10. N Alon et al., 'Adversarial Laws of Large Numbers and Optimal Regret in Online Classification', In Proceedings of the 53rd Annual ACM SIGACT Symposium on Theory of Computing (STOC), pp. 447-455, 2021

11. S Lloyd, 'Least Squares Quantization in PCM', IEEE Trans. Inf. Theory, Vol 28, No 2, pp 129-137, 1982.

12. D Arthur and B Manthey and H Röglin, 'k-Means has Polynomial Smoothed Complexity', arXiv: 0904.1113, 2009.

13. D Sculley, 'Web-Scale K-Means Clustering', Proceedings of the 19th International Conference on World Wide Web, pp 1177-1178, 2010.

14. T Lang, 'AI-Supported Interactive Segmentation of 3D Volumes', PhD thesis, University of Passau, 2021.

15. E Fowlkes and C Mallows, 'A Method for Comparing Two Hierarchical Clusterings', J Am. Stat. Assoc., Vol 78, No 383, pp. 553-569, 1983

16. T Kvalseth, 'Entropy and Correlation: Some Comments', IEEE Trans. Syst. Man Cybern., Vol 17, No 3, pp. 517-519, 1987

17. C Walsh, P Tafforeau and W Wagner et al, 'Imaging intact human organs with local resolution of cellular structures using hierarchical phase-contrast tomography', Nat. Methods, 2021